\documentclass[cameraready]{Interspeech}
\title{The Illusion of Balanced Multimodal Sentiment Analysis: Beyond the Limits of Optimization-Based Methods}

\author[affiliation={1,\dagger, **}, orcid=0009-0002-0485-1083]{Ioanna}{Kaffeza}
\author[affiliation={2,\dagger}, orcid=0000-0002-6042-9584,]{Efthymios}{Georgiou}
\author[affiliation={3,4,5}, orcid=0009-0007-1532-5288]{Alexandros}{Potamianos}
\address{
    $^1$ PERSEE Center, Mines Paris-PSL University, Sophia-Antipolis, France \\
    $^2$ University of Bern, Bern, Switzerland \\
    $^3$ National Technical University of Athens, Athens, Greece \\
    $^4$ Archimedes AI, Athens, Greece \\
    $^5$ Synaptic Bloom PBC, Santa Monica, CA, USA
}

\email{ioanna.kaffeza@minesparis.psl.eu, efthymios.georgiou@unibe.ch, potam@central.ntua.gr}

\keywords{multimodal learning, modality imbalance, optimization dynamics, discriminative evaluation}

\usepackage{comment}

\begin{document}

\maketitle
\begingroup
\renewcommand\thefootnote{\textdagger}
\footnotetext{Work done while at National Technical University of Athens.}

\renewcommand\thefootnote{**}
\footnotetext{Indicates the corresponding author.}
\endgroup
\begin{abstract}
Multimodal Sentiment Analysis (MSA) remains constrained by modality imbalance, yet the field continues to rely on optimization-based balancing methods that promise more than they deliver. We provide three contributions: 1) a unified evaluation framework testing gradient and loss-based balancing strategies under controlled settings; 2) a theoretical diagnosis explaining \textit{why} these methods fail, as they conflate fitting speed with discriminative contribution; and 3) a research agenda toward held-out discriminative modality valuation. Experiments on CMU-MOSI and CMU-MOSEI reveal three shortcomings: no strategy reliably outperforms Late Concatenation; performance is sensitive to hyperparameters; and even ratio calibration fails to yield consistent gains. The core issue is fundamental: loss is not utility, and gradients are not importance. Modality imbalance remains unresolved, motivating utility estimation from held-out performance.

\end{abstract}

\section{Introduction}
Multimodal learning promises a deeper understanding of human affect by combining language, vision, and audio within a single model. These modalities offer complementary signals that should improve sentiment analysis \cite{Poria2017ARO}. Yet empirical findings show that multimodal systems often underperform compared to their unimodal counterparts \cite{wang2020makes}. This counterintuitive mismatch demands a closer examination of how multimodal models are trained.

A key challenge is \emph{modality imbalance}. Each modality exhibits different representational geometry, temporal resolution, information density, and learning dynamics~\cite{georgiou2025multimodal, liang2024foundations}. As a result, one modality tends to dominate optimization, rapidly shaping the shared representation space while the others are undertrained and underrepresented. Joint training under a single objective rarely produces balanced fusion \cite{wang2020makes, wu2022characterizingovercominggreedynature, huang2022modalitycompetitionmakesjoint}. Instead, models overfit to the strongest modality and generalize poorly despite having access to multiple sources of information.

As a remedy, the field has invested in optimization-based interventions. Prior work \cite{wang2020makes, wu2022characterizingovercominggreedynature,peng2022balancedmultimodallearningonthefly,li2023boostingmultimodalmodelperformance} attempts to adjust gradients to compensate for heterogeneous learning rates. Other approaches \cite{fan2022pmrprototypicalmodalrebalance,hua2024reconboostboostingachievemodality} reweight losses in an effort to recalibrate each modality's influence during training. Although these methods differ in implementation, they share a common premise: the assertion that manipulating the optimization process can restore balance among modalities and thereby unlock the full potential of multimodal learning.

This work provides a unified comparative analysis of optimization-based strategies for mitigating modality imbalance in Multimodal Sentiment Analysis. We offer three contributions:

\begin{enumerate}
    \item \textbf{A unified evaluation framework} that tests gradient-based methods (OGM-GE \cite{peng2022balancedmultimodallearningonthefly}, AGM \cite{li2023boostingmultimodalmodelperformance}) alongside loss-based approaches (PMR \cite{fan2022pmrprototypicalmodalrebalance}, ReconBoost \cite{hua2024reconboostboostingachievemodality}) under controlled conditions on CMU-MOSI \cite{zadeh2016mosimultimodalcorpussentiment} and CMU-MOSEI \cite{zadeh2018multimodal}, isolating how each method behaves across different dominance scenarios.
    \item \textbf{A theoretical diagnosis} explaining why these methods fail to consistently outperform baselines. Specifically, optimization-based methods confuse how quickly a modality fits training data with how much it actually contributes to correct predictions; a category error that the field has encountered before. In the 1990's, the audio-visual speech recognition community demonstrated that generative likelihood ratios cannot reliably estimate modality utility, leading to a decisive shift toward discriminative training criteria. The same fundamental misalignment now reappears in modern optimization-based balancing methods.    
    \item \textbf{A research agenda} toward held-out discriminative modality valuation, supported by controlled diagnostic experiments.
\end{enumerate}

\begin{figure}[t]
  \centering
  \includegraphics[width=1\columnwidth]{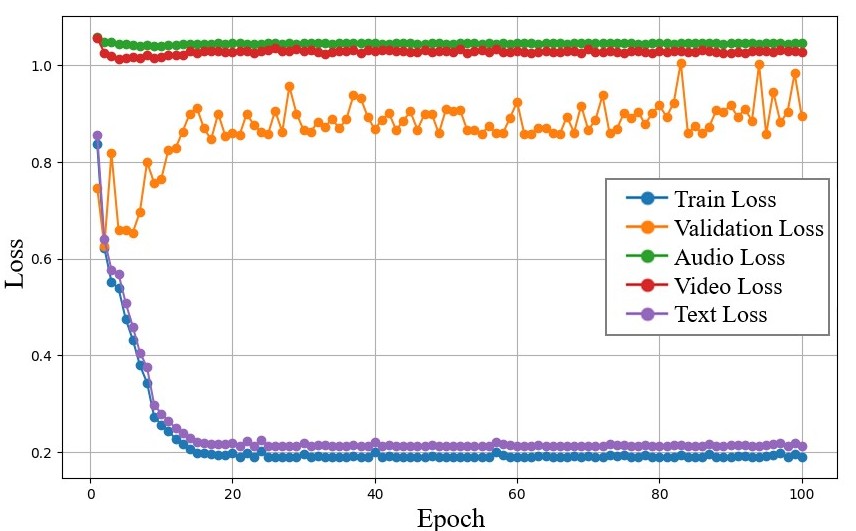} 
  \caption{The Failure of Joint Training.
    Training, validation, and unimodal losses for a representative Late Concatenation run on CMU-MOSI. The text modality rapidly achieves the lowest loss and steers the entire optimization, while audio and visual losses remain largely flat. The multimodal curve mirrors the text-only curve, showing that joint training collapses to the dominant modality. Adapted from \cite{kaffeza2025optimization}.
}
\label{fig:hero}
\end{figure}
We further study how architectural and training factors, including the choice of optimizer, batch size, and the duration of external modulation applied during backpropagation, influence the stability and effectiveness of these techniques. For methods that depend on ratio based estimates of modality contribution \cite{peng2022balancedmultimodallearningonthefly,li2023boostingmultimodalmodelperformance,fan2022pmrprototypicalmodalrebalance}, we introduce a dedicated development set to supply a more stable and unbiased calibration signal, allowing these approaches to be tested under conditions most favorable to their design. 
Taken together, these components allow us to address a central question in multimodal learning: \emph{how effectively can optimization-based reweighting address modality imbalance? Our answer is sobering: not effectively, because these methods measure the wrong signal.}
\vspace{-0.1cm}

\section{Related Work and Background}

\subsection{The Landscape of Multimodal Sentiment Analysis}

Multimodal Sentiment Analysis (MSA) aims to predict emotional and opinion-related states by jointly modeling language, visual, and acoustic cues. Early MSA approaches relied on simple fusion strategies, including early fusion \cite{baltrusaitis2019multimodal} and late fusion \cite{atrey2010multimodal}. Subsequent architectures introduced richer cross-modal reasoning through attention \cite{vaswani2023attentionneed} and large-scale pretraining, such as MulT \cite{tsai2019multimodaltransformerunalignedmultimodal} or self-supervised variants like Self-MM \cite{yu2021learningmodalityspecificrepresentationsselfsupervised}. 

In contrast, we adopt a simpler LSTM-based~\cite{article2} late-fusion architecture to isolate unimodal optimization dynamics without cross-modal overhead. Even with increasingly sophisticated fusion designs and well-established benchmarks, many models appear to operate under the assumption  that all modalities contribute equally during training. This assumption routinely fails \cite{wang2020makes,wu2022characterizingovercominggreedynature,huang2022modalitycompetitionmakesjoint}, exposing a persistent gap between multimodal model design and actual learning behavior.

\begin{table*}[t]
\caption{Comparison of model performance on CMU-MOSI and CMU-MOSEI across three modality combinations. Results are reported as average accuracy (\%) over five runs for the Audio-Video (A-V),
Text-Video (T-V), and Audio-Text-Video (A-T-V) setups. \textit{Late Concatenation} serves as the primary baseline, and each method’s absolute change relative to this baseline ($\Delta$) is shown for the corresponding modality setup. Entries marked ``--'' indicate that the method is not applicable to the three-modality (A-T-V) configuration. Across all settings, $\Delta$ values remain small indicating no consistent improvement over the baseline and suggesting that apparent gains may fall within seed variance. Results reproduced from \cite{kaffeza2025optimization}.}
\label{table:unified-results}
\centering
\footnotesize

\begin{tabular}{lcccccccccccc}
\toprule
\textbf{Method} 
& \multicolumn{4}{c}{\textbf{A--V}} 
& \multicolumn{4}{c}{\textbf{T--V}} 
& \multicolumn{4}{c}{\textbf{A--T--V}} \\
\cmidrule(lr){2-5} \cmidrule(lr){6-9} \cmidrule(lr){10-13}
& MOSI & $\Delta$ & MOSEI & $\Delta$
& MOSI & $\Delta$ & MOSEI & $\Delta$
& MOSI & $\Delta$ & MOSEI & $\Delta$ \\
\midrule

Ensemble
& 47.96 &  & 32.63 &
& 73.35 &  & 43.94 &
& 72.62 &  & 40.40 &  \\
Uni-Pre Finetuned
& 51.46 &  & 32.55 &
& 75.72 &  & 45.22 &
& 75.65 &  & 44.65 &  \\

\midrule
Late Concatenation
& 54.93 &  & 32.55 &
& 74.35 &  & 43.99 &
& 74.87 &  & 44.46 &  \\

\midrule
OGM \cite{peng2022balancedmultimodallearningonthefly}
& 53.30 & -1.63 & 32.67 & +0.12
& 75.04 & +0.69 & 44.15 & +0.16
& -- & -- & -- & -- \\
OGM-GE \cite{peng2022balancedmultimodallearningonthefly}
& 52.48 & -2.45 & 32.40 & -0.15
& 73.50 & -0.85 & 43.58 & -0.41
& -- & -- & -- & -- \\
ACC
& 52.39 & -2.54 & 32.56 & +0.01
& 74.67 & +0.32 & 44.12 & +0.13
& -- & -- & -- & -- \\
AGM \cite{li2023boostingmultimodalmodelperformance}
& 53.73 & -1.20 & 32.71 & +0.16
& 74.61 & +0.26 & 44.15 & +0.16
& -- & -- & -- & -- \\
PMR \cite{fan2022pmrprototypicalmodalrebalance}
& 51.52 & -3.41 & 32.44 & -0.11
& 75.51 & +1.16 & 44.29 & +0.30
& -- & -- & -- & -- \\
ReconBoost \cite{hua2024reconboostboostingachievemodality}
& 47.29 & -7.64 & 33.14 & +0.59
& 74.79 & +0.44 & 44.78 & +0.79
& 75.09 & +0.22 & 44.42 & -0.04 \\

\bottomrule
\end{tabular}
\end{table*}

\subsection{Why Joint Training Favors Certain Modalities}  

Multimodal models rarely learn all modalities at the same pace \cite{wang2020makes}. Differences in representational geometry, temporal resolution, noise, information density and other factors~\cite{georgiou2025multimodal, liang2024foundations} mean that some modalities fit quickly and risk overfitting, while others generalize better but learn more slowly. This mismatch pushes the optimization process to promote the fast-fitting modality and underpin the rest (Figure \ref{fig:hero}). The Greedy Learner Hypothesis \cite{wu2022characterizingovercominggreedynature} formalizes this behavior: multimodal networks tend to favor modalities that yield rapid loss reductions. At the same time, modalities compete for representation capacity \cite{huang2022modalitycompetitionmakesjoint}. The dominant modality drives the entire learning process, leaving the others with minimal impact.

Formally, let $\mathcal{D}_{\text{train}}=\{(x_i,y_i)\}_{i=1}^N$ be a training set with $M$ modalities per input. Each modality $k$ is processed by a feature extractor $F_k(\theta_k)$, producing features $F_k(\theta_k; m_i^k)$ that are concatenated into a joint representation
\begin{equation}
\Phi^M(x_i) = [F_1(\cdot) : \cdots : F_M(\cdot)] .
\end{equation}
which is then passed to a classifier. During backpropagation, parameters for each modality are updated as
\begin{equation}
\theta_k^{t+1} = \theta_k^t - \eta \, \nabla_{\theta_k^t} L(\Phi^M(x), y) .
\end{equation}
with learning rate $\eta$. Although the encoders are separate, late concatenation forces all modalities to share the same loss signal, meaning that the dominant features in $\Phi^M$ steer the overall gradient. By the chain rule, the gradient for modality $k$ decomposes as
\begin{equation}
\nabla_{\theta_k} L =
\frac{\partial L}{\partial \Phi^M}
\cdot
\frac{\partial \Phi^M}{\partial \Phi^k}
\cdot
\frac{\partial \Phi^k}{\partial \theta_k} .
\end{equation}
where $\Phi^k$ is the representation from modality $k$. When the loss gradient aligns more strongly with a particular modality $k^*$, we obtain
\begin{equation}
\lVert \nabla_{\theta_{k^*}} L \rVert \gg \lVert \nabla_{\theta_j} L \rVert
\qquad (j \neq k^*) .
\end{equation}

In practice, this means that the optimization process allocates most of its learning capacity to the dominant modality, while providing only weak updates to the others. This is a structural consequence of joint training under a shared objective, revealing a core limitation of fusion strategies. Suboptimal use of multimodal features creates a serious obstacle in tasks where these modalities carry essential complementary information \cite{wu2022characterizingovercominggreedynature}. Zhang et al. \cite{zhang2024understandingunimodalbiasmultimodal} theoretically show that joint training can induce a transient unimodal phase that results in persistent modality bias, which gradient modulation alone cannot fully correct. As a consequence, the depth and diversity of learned representations are limited \cite{huang2022modalitycompetitionmakesjoint}. This imbalance weakens generalization and robustness \cite{wu2022characterizingovercominggreedynature,xu2023mmcosinemultimodalcosineloss}. In real-world scenarios, leaning too heavily on the dominant modality can make models sensitive to noise or missing information in that modality at test time. 

\subsection{Toward Healing Modality Imbalance}

To counter modality imbalance, some methods select only the most informative modalities during training \cite{georgiou2021m,alfasly2022learnableirrelevantmodalitydropout,JMLR:v25:23-0439}. A second line of work modifies or reweights gradients \cite{wang2020makes,wu2022characterizingovercominggreedynature,peng2022balancedmultimodallearningonthefly,li2023boostingmultimodalmodelperformance,wu2022scalingmultimodalpretrainingcrossmodality,guo2024classifierguidedgradientmodulationenhanced,wei2024mmparetoboostingmultimodallearning} to steer the optimization process toward a more balanced contribution across modalities. A third group adjusts the loss function \cite{fan2022pmrprototypicalmodalrebalance,hua2024reconboostboostingachievemodality,xu2023mmcosinemultimodalcosineloss} to strengthen weaker modalities or temper the influence of dominant ones. In this work, we focus on optimization-based balancing during joint training for Multimodal Sentiment Analysis under a controlled late-fusion setting, rather than explicit modality selection or higher-capacity fusion architectures.

Gradient based methods extend traditional optimizers such as Adam \cite{kingma2017adammethodstochasticoptimization} and Stochastic Gradient Descent (SGD) by assigning adaptive weights to modality specific gradients. OGM-GE \cite{peng2022balancedmultimodallearningonthefly} suppresses gradients from the dominant modality, while AGM \cite{li2023boostingmultimodalmodelperformance} introduces competition free states and real time modulation to reduce interference between modalities.

Loss-based approaches such as PMR \cite{fan2022pmrprototypicalmodalrebalance} and ReconBoost \cite{hua2024reconboostboostingachievemodality} modify the training objective to amplify weaker modalities. PMR employs prototype-based and entropy regularization, while ReconBoost alternates modality-specific updates with reconcilement terms to rebalance their contributions.

Together, these methods attempt to prevent multimodal models from collapsing onto the dominant modality and to ensure that all streams contribute meaningfully. In this work, we evaluate how effectively these strategies address imbalance under a range of controlled conditions.

\subsection{Modality Likelihood Ratios: A Historical Perspective}

The use of likelihood ratios for modality weighting originates in early audio-visual speech recognition. Multi-stream HMM systems showed that generative likelihoods provide unstable estimates of modality reliability, especially under changing noise conditions, and must therefore be combined with confidence measures or discriminative criteria \cite{Potamianos2003RecentAI,articlepotamg,679695,Heckmann2002}. These early approaches are particularly relevant as they frame multimodal fusion as a weighting problem, anticipating the optimization-based balancing strategies studied today. This line of work also emphasized that modality reliability is inherently sample-dependent rather than a fixed global quantity.

Modern multimodal methods reuse the likelihood-based idea through neural surrogates. OGM-GE \cite{peng2022balancedmultimodallearningonthefly} computes modality discrepancies based on modality-specific negative log-likelihoods. AGM \cite{li2023boostingmultimodalmodelperformance} derives Shapley contribution scores from differences in modality-wise negative log-likelihoods. PMR \cite{fan2022pmrprototypicalmodalrebalance} uses class prototypes to approximate class-conditional likelihoods via sample-to-prototype distances, deriving likelihood ratios that guide modality rebalancing. In each case, these likelihood-derived signals are used to adjust modality weights during training, repeating the same approach that failed in the 1990s.

More recent work continues to highlight that likelihood-based estimates do not always track the true, sample-level discriminative contribution of each modality \cite{wei_enhancing_2024}. This persistent mismatch motivates a re-examination of how modality reliability should be estimated in contemporary multimodal learning.

\section{Unified Evaluation Framework}

Using a unified setup with fixed architecture and training, we revisit multimodal optimization in MSA and evaluate two gradient-based and two loss-based approaches that attempt to remedy modality imbalance.

\begin{table}[t]
\caption{Accuracy (\%) of Audio-Video (A-V), Text-Video (T-V), and Audio-Text-Video (A-T-V) models on CMU-MOSI, CMU-MOSEI with ReconBoost under Adam and SGD, showing its sensitivity to optimizer choice. Results reproduced from \cite{kaffeza2025optimization}.}
\label{table:reconboost-results}
\centering
\footnotesize
\setlength{\tabcolsep}{2.2pt}
\renewcommand{\arraystretch}{1.02}

\begin{tabular}{llcccccc}
\toprule
\textbf{Opt.} & \textbf{Method}
& \multicolumn{2}{c}{\textbf{A--V}}
& \multicolumn{2}{c}{\textbf{T--V}}
& \multicolumn{2}{c}{\textbf{A--T--V}} \\
\cmidrule(lr){3-4} \cmidrule(lr){5-6} \cmidrule(lr){7-8}
& & MOSI & MOSEI & MOSI & MOSEI & MOSI & MOSEI \\
\midrule 
Adam  & Baseline   & 54.93 & 32.55 & 74.35 & 43.99 & 74.87 & 44.46 \\
  & ReconBoost & 47.29 & 33.14 & 74.79 & 44.78 & 75.09 & 44.42 \\

\midrule

SGD  & Baseline   & 49.97 & 32.51 & 73.21 & 44.46 & 73.24 & 44.80 \\
  & ReconBoost & 49.88 & 32.42 & 73.45 & 44.37 & 73.07 & 44.30 \\
\bottomrule
\end{tabular}
\end{table}

\subsection{Experimental Setup}

\subsubsection{Optimization Methods Under Evaluation}

\paragraph*{On-the-fly Gradient Modulation (OGM)} \cite{peng2022balancedmultimodallearningonthefly} computes output-likelihood discrepancy ratios to identify the dominant modality and suppresses its gradients to give weaker modalities space to learn. This mechanism assumes that downscaling the fastest-learning modality promotes balance across learning speeds. We also evaluate its Generalization Enhancement (GE)\cite{peng2022balancedmultimodallearningonthefly} extension, which adds a regularizer for improved robustness, and a variant (ACC) that instead amplifies the weakest modality based on the canonical discrepancy ratio.

\paragraph*{Adaptive Gradient Modulation (AGM)} \cite{li2023boostingmultimodalmodelperformance} uses Shapley-inspired mono-modal outputs to estimate how much each modality contributes to the prediction. These estimates are computed from differences in modality-specific negative log-likelihoods, producing modulation coefficients that reshape gradients under the assumption that contribution-aware updates lead to fairer optimization.

\paragraph*{Prototypical Modal Rebalance (PMR)} \cite{fan2022pmrprototypicalmodalrebalance} builds class-specific prototypes and reweights the loss according to an imbalance ratio derived from distances to these prototypes. We exclude PMR's entropy regularizer to focus strictly on the penalization effect and to maintain comparability with OGM and AGM methods.

\paragraph*{ReconBoost} \cite{hua2024reconboostboostingachievemodality} alternates updates across unimodal encoders and adopts Kullback--Leibler (KL)-based regularization \cite{Kullback1951OnIA} to encourage each modality to correct errors made by the others, as measured during training. Its central assumption is that alternating updates combined with cross-modal reconstruction pressure yields more balanced learning dynamics.

\subsubsection{Model Architecture}

Each modality is encoded with a unidirectional LSTM; representations are concatenated and fed to a shared classifier. We deliberately adopt a simple late-fusion architecture to isolate optimization dynamics from architectural confounds, ensuring that any observed effects reflect training-signal behavior rather than representational capacity. The architecture is fixed across methods and trained from scratch to isolate optimization effects from architectural variation. In addition to the optimization methods, we evaluate three standard training strategies: (1) soft-voting ensembles, averaging independent unimodal predictions; (2) uni-modality pre-finetuning, where encoders are trained individually before joint late-fusion finetuning; and (3) joint late concatenation without balancing. These baselines provide a reference for interpreting balancing methods.

\subsubsection{Training Configuration Sensitivity Analysis}

We evaluate generalization by varying core training choices, such as optimizer, training duration, and modulation period, which influence convergence and modality dynamics. These controlled variations reveal how much behavior is driven by training configuration rather than method design.

To investigate how robust modality estimates are, we compute modality-specific ratios on a separate development set rather than on the training batch. Using training data alone ties these estimates to batch size and introduces instability. Decoupling ratio computation from the training loop yields a more stable, unbiased, and generalizable signal of modality importance. We apply this modification only when it aligns with a method's original design \cite{peng2022balancedmultimodallearningonthefly,li2023boostingmultimodalmodelperformance,fan2022pmrprototypicalmodalrebalance}. This modification represents a partial move toward discriminative estimation: the development set provides signal about generalization rather than training fit.

To expose how each optimization method behaves under different dominance conditions, we construct three controlled imbalance setups: two weaker modalities (Audio-Video); one dominant and one weak modality (Text-Video), while omitting Text-Audio
due to its analogous text-dominated behavior; and a fully imbalanced case where all three modalities are present. These scenarios let us test whether the methods genuinely adapt to changing modality strengths rather than relying on favorable conditions. We characterize modality balance through performance stability and dominance trajectories across controlled imbalance settings.

\subsubsection{Datasets and Feature Representation}

We evaluate on CMU-MOSI \cite{zadeh2016mosimultimodalcorpussentiment} and CMU-MOSEI \cite{zadeh2018multimodal} using standard features: 768-dim BERT \cite{devlin2019bert} for text, COVAREP \cite{6853739} for audio, and OpenFace \cite{7477553} for vision. Fixed representations isolate optimization effects. A small held-out development set (100 samples for CMU-MOSI and 200 for CMU-MOSEI) is used for auxiliary metrics. Sentiment is treated as classification: three classes for MOSI and seven ordinal classes for MOSEI following \cite{Delbrouck_2020}.

\subsubsection{Training Protocol}

All models follow the same training protocol. We use Adam \cite{kingma2017adammethodstochasticoptimization} with a \texttt{ReduceLROnPlateau} scheduler (factor 0.1), batch sizes of 16 (MOSI) and 32 (MOSEI) \cite{yu2021learningmodalityspecificrepresentationsselfsupervised, wu2024multimodalmultilossfusionnetwork}, and early stopping (patience 8). Cross-entropy loss is used throughout. Because optimization strategies are sensitive to learning rate, each method receives its own grid search; all other settings are fixed. We report the best configuration per method, averaged over five seeds, selected by validation loss. Experiments run on a single NVIDIA GTX~1080~Ti (12GB).

\subsection{Empirical Results}

These methods aim to balance modalities, but empirical results reveal a more nuanced picture. 

\subsubsection{Strong Baselines, Limited Gains}

Table \ref{table:unified-results} shows a consistent trend: strong baselines remain difficult to surpass. On CMU-MOSI, Late Concatenation achieves the highest Audio-Vision accuracy, with OGM and AGM matching it only sporadically. On CMU-MOSEI, AGM offers a slight gain. In Text-Vision and trimodal settings, Uni-Pre Finetuned is the most reliable performer, outperforming all optimization-based methods across datasets. Adding audio yields minimal benefit, especially on CMU-MOSEI, reaffirming the dominance of text and the limited corrective power of reweighting strategies. Overall, the evaluated methods provide only modest improvements, rarely exceeding the robustness of the baselines they aim to improve.

\subsubsection{Adam Does the Heavy Lifting}

\begin{figure}[t]
  \centering

  \includegraphics[width=0.95\columnwidth]{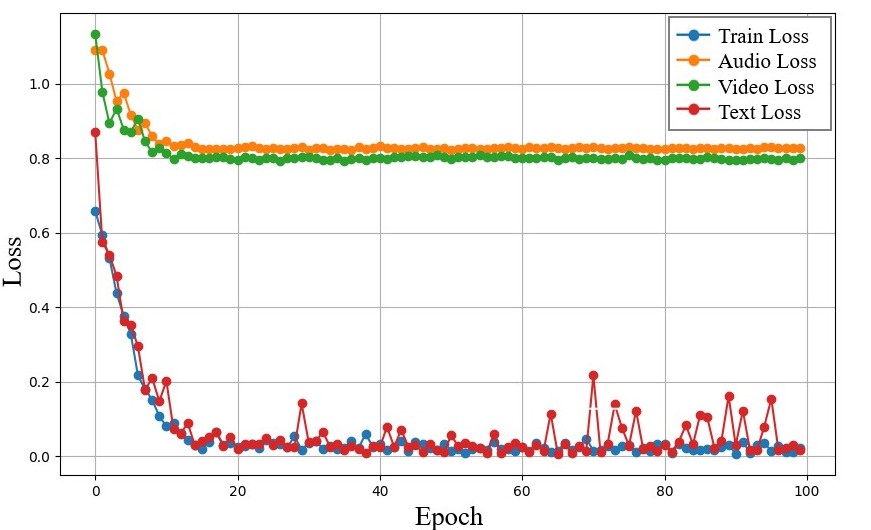}
  \includegraphics[width=0.95\columnwidth]{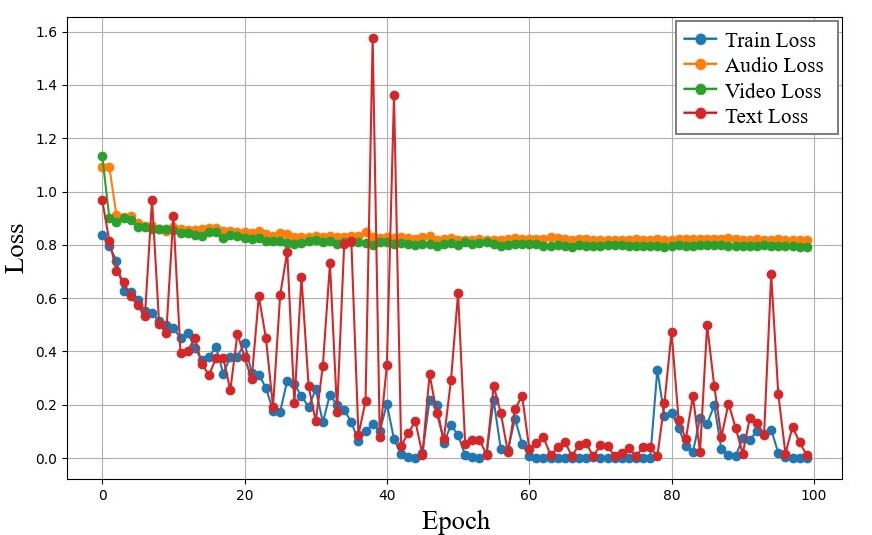}

  \caption{Training losses for ReconBoost on CMU-MOSI with Adam (top) and SGD (bottom). Text remains dominant, while audio and video losses drop in the early phase, showing that modulation is only useful briefly. SGD is more volatile than Adam but converges to the same pattern. Reproduced from \cite{kaffeza2025optimization}.}
  \label{fig:modality_loss_reconboost}
\end{figure}

Our experiments show that Adam generally outperforms SGD across most methods and datasets (Table \ref{table:reconboost-results}). Its adaptive learning rates provide the stability multimodal training depends on, while SGD converges slowly and is far more sensitive to hyperparameters (Figure \ref{fig:modality_loss_reconboost}). 

This finding is revealing: the ``balancing'' these methods provide is largely an artifact of favorable optimizer dynamics. When the optimizer changes, the improvements vanish. True modality balancing should be robust to such configuration choices.
\begin{table}[t]
\caption{Accuracy (\%) of Audio-Video (A-V) and Text-Video (T-V) under different modulation schedules. The minimal differences across
schedules indicate that models converge early, making extended modulation unnecessary. Results from \cite{kaffeza2025optimization}.}
\label{table:ogm-modulation-results}
\centering
\footnotesize
\setlength{\tabcolsep}{2.2pt}
\renewcommand{\arraystretch}{1.05}

\begin{tabular*}{\columnwidth}{@{\extracolsep{\fill}}lllcc}
\toprule
\textbf{Dataset} & \textbf{Schedule} & \textbf{Method} & \textbf{A--V} & \textbf{T--V} \\
\midrule

MOSI& 100 epochs & Baseline & 53.09 & 72.68 \\
    & & OGM    & 52.95 & 72.97 \\
    & & OGM-GE & 50.47 & 73.59 \\

\midrule

MOSEI& 5 epochs & Baseline & 32.55 & 43.99 \\
     & & OGM    & 32.67 & 44.15 \\
     & & OGM-GE & 32.40 & 43.58 \\

\midrule
MOSEI  & All epochs & Baseline & 32.55 & 43.99 \\
  & & OGM    & 32.59 & 44.02 \\
  & & OGM-GE & 32.43 & 43.13 \\

\bottomrule
\end{tabular*}
\end{table}
\subsubsection{Long Training, Little Gain}

Our experiments show that OGM and OGM-GE gain little from extended training or prolonged modulation schedules (Table \ref{table:ogm-modulation-results}). Although OGM-GE is designed to benefit from long training due to its Gaussian noise component, it still performs below strong baselines. In practice, these models converge within the early epochs, making additional modulation largely redundant.

This pattern aligns with the theoretical analysis of Zhang et al. \cite{zhang2024understandingunimodalbiasmultimodal}: unimodal bias is established early and becomes permanent through overfitting. Gradient modulation applied after this critical period cannot reverse the damage.

\subsubsection{Development Set Helps, but Not Enough}

\begin{figure*}[t]
  \centering

  % First image
  \begin{minipage}[b]{0.32\textwidth}
    \centering
    \includegraphics[width=\linewidth]{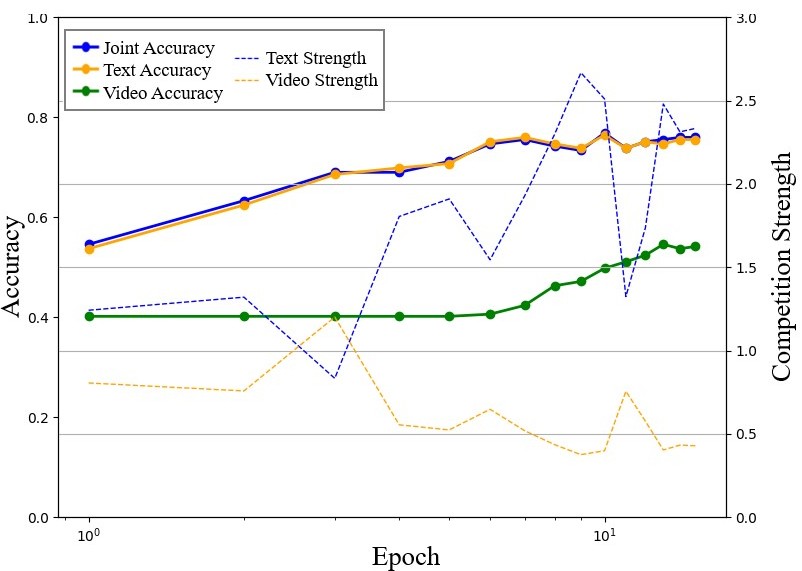}
    \caption*{(a) Late Concatenation}
  \end{minipage}
  \hfill
  % Second image
  \begin{minipage}[b]{0.32\textwidth}
    \centering
    \includegraphics[width=\linewidth]{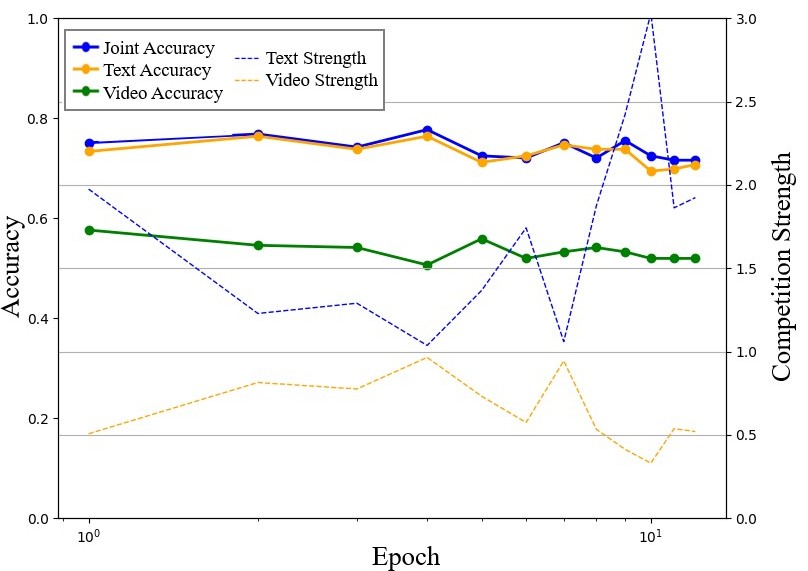}
    \caption*{(b) AGM}
  \end{minipage}
  \hfill
  % Third image
  \begin{minipage}[b]{0.32\textwidth}
    \centering
    \includegraphics[width=\linewidth]{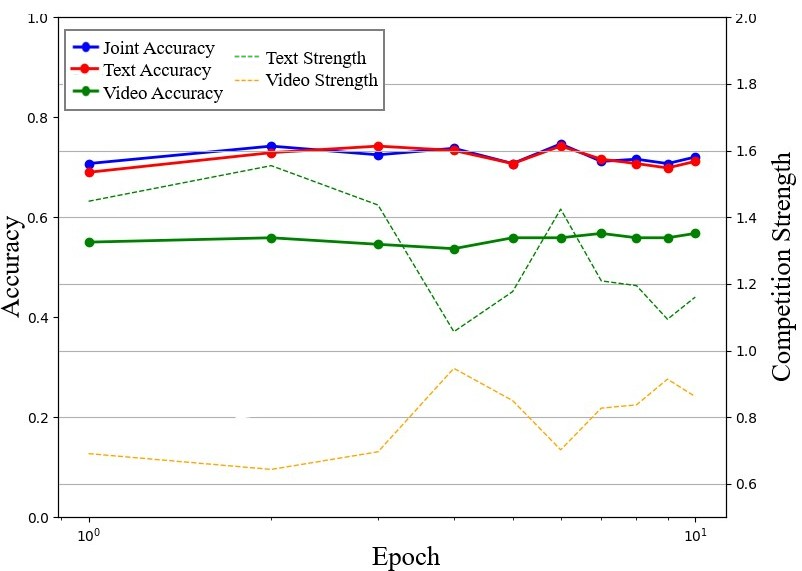}
    \caption*{(c) AGM with development set}
  \end{minipage}

  \caption{Baseline (a), AGM (b), and AGM with development set (c). Training-only estimates lead to unstable modality strengths, while development-set calibration results in smoother, more consistent trajectories. Reproduced from \cite{kaffeza2025optimization}.}
  \label{fig:agm_ratios}
\end{figure*}

Using a development set to compute discrepancy (OGM-GE), strength (AGM), or imbalance (PMR) produces more stable and less biased estimates than relying on the training batch. This decoupling makes the methods less sensitive to batch size and better reflects true modality contributions. As shown in Figure \ref{fig:agm_ratios}, AGM calibrated with a development set (Figure \ref{fig:agm_ratios}(c)) tempers the dominance of text and strengthens the video modality (Figure \ref{fig:agm_ratios}(a)), leading to a more balanced contribution profile compared to standard AGM (Figure \ref{fig:agm_ratios}(b)). Overall, development set calibration stabilizes these methods, even if it does not fundamentally change their performance limits.

This is the most instructive result: moving toward held-out estimation helps, but using \emph{losses} rather than held-out \emph{discriminative performance} reveals the fundamental limitation. The development set provides a better estimate of generalization loss, but loss alone remains a poor indicator.

\subsection{Why These Methods Fall Short: The Emerging Picture}

Our results reveal a consistent pattern across all methods. When we change the optimizer, the improvements vanish, showing that the methods rely on configuration choices rather than true modality balancing. Their modality estimates are too noisy to be trusted during training, so they rely on a development set to correct their decisions. Longer training or extended modulation also offers no advantage, since models converge early and additional gradient manipulation does not alter the imbalance. 

The combination of high variance, sensitivity to configuration choices, and small gains that vanish under mild changes mirrors the conclusions of other studies \cite{wei_enhancing_2024, kontras2024multimodalfusionbalancinggametheoretic,wei_diagnosing_2024}, which report similarly fragile behavior across datasets, architectures, and imbalance scenarios. Taken together, the evidence indicates that these approaches do not correct the underlying optimization dynamics; they simply shift where the instability becomes visible.

We therefore turn to a deeper analysis of the mechanisms behind this behavior, asking whether the issue reflects a more fundamental limitation of training-time optimization signals as indicators of modality contribution.

\section{The Fundamental Problem: 
Fitting is not Contributing}

Losses, gradients, and likelihood ratios measure how quickly a modality \emph{fits} the training data, not how much it \emph{contributes} to correct predictions at test time. This distinction is central to understanding the limitations of optimization-based balancing. It is supported by both theoretical arguments and historical evidence, and motivates diagnostics experiments as well as a principled path forward. The goal of this section is to diagnose the structural mismatch between fitting signals and discriminative utility, and to outline principles for future method development.

\subsection{A Theoretical Diagnosis}

\subsubsection{Loss Measures Fitting Speed, Not Utility}

A modality can achieve low training loss through several mechanisms: 1) it carries genuinely strong discriminative cues, 2) it memorizes/overfits training examples, or 3) it picks spurious correlations (shortcut learning) present in the training set. 
Loss alone cannot distinguish between these cases \cite{zhang2017understandingdeeplearningrequires}. A dominant modality, like text in MSA~\cite{georgiou2021m, hazarika-etal-2022-analyzing}, may legitimately achieve lower loss because it is more informative; yet the same low-loss signal can also identify a modality that has simply learned or overfit faster. 

This ambiguity undermines the core logic of loss-based reweighting. When PMR \cite{fan2022pmrprototypicalmodalrebalance} penalizes modalities with smaller prototype distances, or when ReconBoost \cite{hua2024reconboostboostingachievemodality} alternates updates based on loss trajectories, these methods cannot determine whether they are suppressing an overfit modality or handicapping a genuinely informative one.

\subsubsection{Gradients Measure Rate of Change, Not Importance}

Gradient-based methods face a parallel problem. Large gradients indicate that a modality's parameters are changing rapidly-but rapid change could mean the modality is learning useful features, or that it is oscillating due to noise, or that it is in an early phase of training that will later stabilize. Small gradients are equally ambiguous: a modality may have converged to a good representation, or it may be stuck in a poor local minimum, or it may be receiving suppressed updates due to dominance by another modality \cite{ zhang2024understandingunimodalbiasmultimodal, adebayo2020sanitycheckssaliencymaps}.

OGM-GE \cite{peng2022balancedmultimodallearningonthefly} suppresses gradients from the ``dominant'' modality identified by output likelihood discrepancies, and AGM \cite{li2023boostingmultimodalmodelperformance} modulates gradients based on Shapley-inspired contribution estimates. Both methods assume that gradient magnitude correlates with modality importance. Yet gradient dynamics reflect optimization trajectories, not discriminative value.

\subsubsection{Likelihood Ratios Inherit the Same Flaw}

Modern methods that compute modality “contributions” from negative log-likelihoods or softmax outputs are neural analogues of the generative likelihood-ratio approaches used in 1990s audio-visual speech recognition. The same limitation identified then still applies: generative likelihoods measure how well each modality explains the data, not how much it improves class discrimination \cite{Potamianos2003RecentAI,679695,Heckmann2002}.

As shown in \cite{679695}, learning stream weights discriminatively, by directly minimizing classification error, succeeds where likelihood-based weighting fails \cite{679695}. The key insight is that modality reliability must be inferred from discriminative performance rather than generative fit, an idea largely overlooked in modern multimodal learning.

\subsubsection{Why Complementarity Cannot Be Estimated from Loss}

The Generalized Ambiguity Decomposition (GAD) theorem \cite{audhkhasi2013gad} shows that for an ensemble (analogous to modality-specific encoders),
\begin{equation}
\text{Ensemble Loss} \approx \overline{\text{Expert Loss}} - \text{Diversity},
\end{equation}
where $\overline{\text{Expert Loss}}$ is the average loss of individual experts and Diversity measures disagreement among them. Fusion gains thus arise from the complementarity factor within Diversity, not from individual losses alone. A modality with higher marginal loss may still add value by correcting others’ errors.

Since marginal loss contains no information about error overlap, complementarity cannot be inferred from it. Modalities with identical losses may contribute very differently depending on whether their errors coincide or diverge. Loss-driven reweighting therefore risks miscalculating modality contribution unless cross-modal interactions are modeled.

\subsubsection{Sample-Level Variation Defeats Global Reweighting}

Even if loss or gradient signals were accurate indicators of modality utility, they would still fail because modality contributions vary at the sample level \cite{Potamianos2003RecentAI,articlepotamg,wei_enhancing_2024}. For some samples, audio may be more informative; for others, video may dominate; for still others, only the combination provides discriminative signal. Global reweighting schemes, whether applied per-batch or per-epoch, cannot capture this variation.

\subsubsection{Summary: The Category Error}
The methods we evaluate commit a category error: they use \emph{fitting signals} (loss, gradients, likelihoods) as proxies for \emph{contribution signals} (discriminative value, complementarity, test-time utility). This error has been identified before and the solution was clear: use discriminative criteria estimated from held-out performance, not generative signals from training.

\subsection{A Minimal Diagnostic Result}
We now formalize modality contribution from a discriminative perspective and establish a minimal result that motivates it.
\subsubsection{Controlled Validation: XOR-Gated Complementarity}

We construct a synthetic experiment with known (ground-truth) complementarity.

\textbf{Setup.} In the \texttt{LOW} condition, modality A carries the label signal; B is pure noise. In the \texttt{HIGH} condition, a latent variable $z$ determines which modality is informative per-sample, but $z$ is encoded via XOR across both modalities (A contains random bit $r$; B contains $r \oplus z$). Neither modality alone reveals $z$. Fusion is necessary to uncover it.

\begin{table}[h]
\centering
\caption{Classification accuracy on the synthetic complementarity task. Columns report test accuracy for unimodal models (A-only, B-only), a late-fusion model using both modalities (Fusion), and the same fusion model trained with OGM (Fusion+OGM). In the HIGH condition, OGM degrades fusion performance.}
\begin{tabular}{lcccc}
\toprule
\textbf{Condition} & \textbf{A-only} & \textbf{B-only} & \textbf{Fusion} & \textbf{Fusion+OGM} \\
\midrule
\texttt{LOW}  & 100\% & 53\% & 100\% & 100\% \\
\texttt{HIGH} & 49\%  & 54\% & 88\%  & 76\%  \\
\bottomrule
\end{tabular}
\label{tab:xor_results}
\end{table}

\textbf{Results.} In the \texttt{HIGH} condition, OGM \textit{degrades} performance (Table \ref{tab:xor_results}). Both modalities show identical training losses ($\approx 0.69$) and gradient norms throughout, so OGM applies symmetric modulation, disrupting the joint learning required to decode the XOR gate. The method cannot detect that neither modality alone is informative while both together are essential.

This confirms that optimization-based methods cannot estimate complementarity from training-time signals when sample-level variation is high.

\subsubsection{Learning Speed Is Not Utility: A Controlled Demonstration}

Our central hypothesis is that gradient-based measures reflect \emph{optimization speed}, not \emph{test-time utility}. We examine whether gradient magnitude can distinguish genuinely informative fast learning from fast but non-generalizable learning. 

\textbf{Setup.}
We construct a two-modality synthetic classification task with a late-fusion architecture. The model, optimization procedure, and hyperparameters are identical across conditions; only the data-generating process changes. In Condition A, the fast modality provides a strong signal that generalizes to test time, while the slow modality is weaker but informative. In Condition B, the fast modality is deliberately engineered to be perfectly predictive during training through label-dependent prototypes, but this correlation is broken at test time, making it non-generalizable. The slow modality remains informative in both splits. Condition B therefore serves as a controlled setting in which rapid training dynamics arise from spurious structure rather than true signal. We compare standard training (baseline) with OGM-GE modulation.
\begin{table}[t]
\centering
\caption{Controlled experiment comparing a genuinely informative fast modality (Condition A) with a spurious fast learner (Condition B). We report final training and test accuracy for the baseline and OGM-GE. “Fast coeff.” is the epoch-averaged OGM-GE gradient scaling factor applied to the fast modality (values $ \ll 1 $ indicate strong suppression).}
\footnotesize
\setlength{\tabcolsep}{2.2pt}
\renewcommand{\arraystretch}{1.05}
\label{tab:exp2_results}
\begin{tabular}{lcccc}
\toprule
\textbf{Condition} & \textbf{Method} & \textbf{Train Acc.} & \textbf{Test Acc.} & \textbf{Fast Coeff. (avg)} \\
\midrule
A 
& Baseline & 100\% & 99.2\% & 1.00 \\
& OGM-GE   & 100\% & 97.4\% & $\ll 1$ \\[4pt]

B  
& Baseline & 100\% & 10.1\% & 1.00 \\
& OGM-GE   & 100\% & 10.1\% & $\ll 1$ \\
\bottomrule
\end{tabular}
\end{table}

\textbf{Results.}
In both conditions, OGM-GE consistently identifies the fast modality as dominant and suppresses it during training (Table \ref{tab:exp2_results}). However, the implications differ. In Condition A, suppression reduces test accuracy because the fast modality carries genuine signal. In Condition B, suppression is appropriate since the fast modality represents a shortcut. Importantly, the training-time dominance signal is similar in both cases.

These results demonstrate that gradient magnitude captures how quickly a modality fits the training data, but not whether that fit corresponds to robust, generalizable structure. Gradient-based dominance measures therefore conflate fast learning with true informativeness.

\subsubsection{Validation-Based Weight Selection as a Diagnostic}

We evaluate whether global reweighting is structurally sufficient on real data, and whether training-time proxy signals provide a reliable basis for fusion.

\textbf{Setup.}
Encoders are first trained with standard late concatenation without modality reweighting and then frozen. Fusion is analyzed post hoc by reweighting the modality-specific logit contributions of the frozen classifier using scalar weights selected from a small grid of candidate values.

We compare four strategies for selecting a single global fusion weight: 
\texttt{Proxy}, which derives weights from unimodal training cross-entropy; 
\texttt{ValSel}, which selects the weight maximizing validation accuracy; 
\texttt{Oracle}, which selects the best global weight using test labels and therefore represents an upper bound; 
and \texttt{OGM-GE} \cite{peng2022balancedmultimodallearningonthefly}, a representative training-time gradient modulation method.

To examine whether a single global weight adequately reflects modality utility, we also compute a \emph{per-sample optimal weight}. For each sample, we evaluate all candidate fusion weights and select the one that minimizes the cross-entropy loss for that prediction. We summarize the variability of these per-sample optimal weights using two statistics: 
\texttt{Entropy}, which measures the diversity of optimal weights across samples, and 
\texttt{Optimal Weight Disagreement (OWD)}, defined as the percentage of samples whose per-sample optimal weight differs from the validation-selected global weight. This analysis characterizes per-sample optimal fusion weights and how global weighting departs from sample-specific utility.

\textbf{Results.}
Validation-based selection performs comparably to the proxy and remains within 0.3--0.4\% of oracle performance (Table~\ref{tab:val_weight_poc}), indicating that an appropriately chosen global fusion weight captures most aggregate accuracy.

Low sample-level variation would imply that a single global fusion weight suffices across inputs. However, Table~\ref{tab:val_weight_poc} shows consistently non-trivial entropy and high disagreement in most modality combinations, indicating that the per-sample optimal fusion weight frequently differs from the global selection. These results suggest that, while global weighting is adequate in aggregate performance, substantial sample-level variation in optimal weights persists in the evaluated settings.

\begin{table}[t]
\caption{
Acc-2 of global fusion-weight selection strategies on CMU-MOSI and CMU-MOSEI for Audio-Video (A-V), Text-Video (T-V), and Audio-Text-Video (A-T-V). 
Proxy derives weights from unimodal training cross-entropy, ValSel selects the validation-optimal weight, Oracle reports the best global weight using test labels, and OGM-GE corresponds to training-time gradient modulation. 
Ent. denotes the entropy of per-sample optimal weights, and OWD (\%) the Optimal Weight Disagreement with the validation-selected global weight.
}
\label{tab:val_weight_poc}
\centering
\footnotesize
\setlength{\tabcolsep}{4pt}
\renewcommand{\arraystretch}{1.05}
\begin{tabular}{lcccccc}
\toprule
\textbf{Setup} & \textbf{Proxy} & \textbf{ValSel} & \textbf{Oracle} & \textbf{OGM-GE} & \textbf{Ent.} &  \textbf{OWD.(\%)} \\
\midrule
\multicolumn{7}{c}{\textbf{MOSI}} \\
\midrule
A-V  & 0.4223 & 0.4223 & 0.4223 & 0.4223 & 0.681 &  42 \\
T-V  & 0.8003 & 0.7988 & 0.8018 & 0.7851 & 0.510 &  100 \\
A-T-V & 0.7957 & 0.7957 & 0.7988 & --     & 0.649 &  100 \\
\midrule
\multicolumn{7}{c}{\textbf{MOSEI}} \\
\midrule
A-V  & 0.6285 & 0.6285 & 0.6285 & 0.6293 & 0.671 &  39.7 \\
T-V  & 0.8360 & 0.8343 & 0.8343 & 0.8360 & 0.446 & 16.4 \\
A-T-V & 0.8145 & 0.8154 & 0.8178 & --     & 0.555 &  100 \\
\bottomrule
\end{tabular}
\end{table}

\subsubsection{Structural Limits of Global Reweighting}

While Section~4.2.3 shows that global weighting rarely aligns with per-sample optimality, we next quantify the magnitude of this structural limitation.

\textbf{Setup.}
Using the same frozen encoders as in Section 4.2.3, we evaluate the structural limits of global reweighting. 
We compare two upper bounds. The \texttt{Global Oracle} selects a single fusion weight vector that maximizes accuracy on the test set and applies it to all samples. The \texttt{Restricted Per-Sample Oracle} selects, for each test sample independently, the fusion weight vector that yields the correct prediction among the available non-unimodal fusion combinations. 
This oracle therefore represents the best performance achievable if fusion weights could be chosen per sample rather than globally.

\textbf{Results.}
The restricted per-sample oracle achieves 88.57\% accuracy on MOSI and 89.57\% on MOSEI, compared to 79.88\% and 81.78\% for the respective global oracles (Table~\ref{tab:headroom}). This yields structural headroom of +8.7\% and +7.79\%, quantifying the performance unavailable to any method constrained to a single global fusion weight.

\begin{table}[t]
\caption{Accuracy (Acc-2) illustrating the limits of global fusion weighting on CMU-MOSI and CMU-MOSEI (A-T-V). The Global Oracle selects a single fusion weight vector maximizing test accuracy, while the Restricted Per-Sample Oracle selects optimal fusion weights independently for each sample. The gap (Structural Headroom) represents performance unavailable to any method constrained to a single global weight.}
\label{tab:headroom}
\centering
\small
\setlength{\tabcolsep}{4pt}
\renewcommand{\arraystretch}{1.05}
\begin{tabular}{lcc}
\toprule
\textbf{Method} & \textbf{MOSI} & \textbf{MOSEI}\\
\midrule
Global Oracle          & 0.7988 & 0.8178\\
Restricted Per-Sample Oracle & 0.8857 & 0.8957\\
\midrule
\textit{Structural Headroom}
  & +8.70\% & +7.79\%\\
\bottomrule
\end{tabular}
\end{table}

This result makes the limitation of global reweighting concrete:
even a perfect global reweighting strategy—one that always selects
the best possible single weight vector—leaves a 8.7\% accuracy gap
relative to what sample-level weight selection could theoretically achieve. Closing this gap requires methods capable of estimating modality
utility at the sample level rather than relying on a fixed global weighting.

\subsection{A Discriminative Path Forward: Implication and Open Problems}

Optimization-based balancing conflates fitting with contributing. The empirical results in Section~3 and the controlled demonstrations in Section~4.2 suggest that estimating modality utility from held-out discriminative performance is a promising direction, and they raise open questions about how to do so reliably under sample-level variation and complementarity.

\subsubsection{Estimate Utility from Held-Out Performance}

The optimization of modality encoders (which should be trained on training data) must be separated from the optimization of fusion weights (which should be learned from validation discriminative performance). This separation mirrors the logic of Neural Architecture Search, where validation performance guides architecture selection rather than training loss.

Current methods attempt to estimate modality contribution from within the training loop, using signals that reflect fitting rather than utility. The fundamental insight from discriminative stream weighting \cite{679695} is that modality reliability cannot be estimated from generative signals---it must be measured by discriminative performance on held-out data.

\subsubsection{When Should We Expect Training-Time Weighting to Work?}

The preceding analysis in Section~4.2 identifies regimes in which optimization-based weighting may be harmless, though not necessarily helpful. 

\textbf{Low complementarity.} When one modality strictly dominates and others add no complementary information, suppressing weak modalities is unlikely to hurt. But it is also unlikely to help, since the baseline already ignores them.
    
\textbf{Dominance alignment.} When the fast-learning modality is also the most informative at test time, training-time signals accidentally correlate with utility. In sentiment analysis, where text typically dominates, we observe the strongest performance in text-heavy configurations.

\textbf{Low sample-level variation.} When optimal fusion weights are approximately constant across samples, global reweighting may suffice. But this condition rarely holds in practice.

These are alignment regimes rather than guarantees. When complementarity is substantial, when fast learners overfit, or when modality utility varies across samples, the mismatch between fitting and contributing becomes decisive. Characterizing these regimes more formally remains an open problem.

\subsubsection{Open Research Directions}

The diagnostic experiments above make future directions concrete: the structural headroom quantifies the necessity of sample-level valuation; the validation-based selection results demonstrate the feasibility of held-out fusion optimization; the observed disagreement across settings motivates discriminative meta-classification; and the dominance-alignment regimes highlight the importance of robustness-aware modality profiling. Together, these observations reinforce the need to ground modality valuation in held-out discriminative performance:

\textbf{Discriminative Meta-Classification.} Reframe modality balancing as a meta-regression problem solved through discriminative profiling on held-out validation data. Rather than manipulating training-time signals, exhaustively measure which modality combinations actually predict correctly, for each validation sample individually. Train meta-regressors to predict fusion weights from test-time-observable features (modality encodings, prediction confidences, cross-modal agreement), with regularization toward validation-optimized global weights.

\textbf{Sample-Level Modality Valuation.} Building on Wei et al. \cite{wei_enhancing_2024}, develop methods that estimate per-sample modality contributions from validation performance rather than training signals. The key insight is that modality utility varies by sample; global reweighting cannot capture this variation.

\textbf{Robustness Profiling.} Measure modality utility under corruption and noise on validation data. A modality that remains accurate under perturbation provides more robust signal than one that degrades. This robustness information cannot be extracted from clean training loss.

\textbf{Validation-Based Fusion Weight Optimization.} Directly optimize fusion weights to maximize validation accuracy, treating encoder parameters as fixed after initial training. This fully separates encoder optimization from fusion optimization.

\section{Conclusions}

In this work, we conducted a unified evaluation of recent optimization-based methods for modality balancing in Multimodal Sentiment Analysis. We found that no method reliably outperforms simple late fusion. Their core limitation is clear: they treat losses and gradients as indicators of discriminative value.

Modality imbalance persists, text continues to dominate, and performance remains highly sensitive to training choices. Resolving modality imbalance requires moving beyond training loop interventions toward methods that estimate modality value from held-out discriminative performance, a principle established in discriminative stream weighting \cite{679695} and supported by the GAD characterization of diversity \cite{audhkhasi2013gad}.

Our work provides both a diagnostic foundation, establishing what current optimization methods cannot achieve and why, and a research agenda pointing toward discriminative meta-learning frameworks that separate encoder optimization from fusion weight optimization. The path forward is not better gradient modulation or smarter loss reweighting; it is moving beyond purely training-time fitting signals toward held-out, discriminative criteria for modality valuation. The discriminative directions outlined here are the subject of ongoing work.

\section{Generative AI Use Disclosure}
Generative AI tools were used to assist in refining parts of the manuscript text. All scientific content, experimental design, analyses, and conclusions were developed, verified, and approved by the authors, who take full responsibility for the final manuscript.

\bibliographystyle{IEEEtran}
\bibliography{backup}

\end{document}